# Distributed control and navigation system for quadrotor UAVs in GPS-denied environments

Konstantin Yakovlev, Vsevolod Khithov, Maxim Loginov, Alexander Petrov

Institute for Systems Analysis of Russian Academy of Sciences, Moscow, Russia
yakovlev@isa.ru
Soloviev Rybinsk State Aviation Technical University, Rybinsk, Russia
vskhitkov@gmail.com
Soloviev Rybinsk State Aviation Technical University, Rybinsk, Russia
lunarstrainut@gmail.com
NPP SATEK plus, Rybinsk, Russia
gmdidro@gmail.com

**Abstract.** The problem of developing distributed control and navigation system for quadrotor UAVs operating in GPS-denied environments is addressed in the paper. Cooperative navigation, marker detection and mapping task solved by a team of multiple unmanned aerial vehicles is chosen as demo example. Developed intelligent control system complies with on 4D\RCS reference model and its implementation is based on ROS framework. Custom implementation of EKF-based map building algorithm is used to solve marker detection and map building task.

## 1 Introduction

Modern intelligent systems are characterized by increasing autonomy and distribution. From one side autonomy increases in a sense of behavior independence from operator and from other side, in case of multiagent system, its autonomy could be characterized by degree of automation in agents collaboration. Modern systems become distributed both in terms of ensuring a collective interaction of several intelligent agents (the distribution of knowledge and information between agents), and in terms of a distributed software technologies.

We present distributed control and navigation system for unmanned aerial vehicles (UAV), e.g. quadrotors, in GPS-denied environments. We demonstrate architecture and implementation of our system on map building task solved by the collaboration of UAVs. The task is following – N quadrotors (we use AR.Drones in our experiments) are operating in a GPS-denied environment, e.g. a room, and are remotely controlled (via wi-fi link) by software control system on ground station. There are several markers, that placed somewhere in the room (we use cubes with QR-codes on their sides) and AR.Drones should firstly detect markers then compute markers' position relative to drone's start position (individual map) and after that integrate and elaborate shared map.

In this paper first we discuss conceptual framework for layered representation of intelligent system. Next we will focus on marker detection, drone and marker position estimation and map building algorithms. Next we describe control system software implementation based on ROS framework [1]. In conclusion we present experimental evaluation results and future direction of our work.

### 1.1 Related works

Previous research in distributed robot control system includes: [2] which propose ALLIANCE fault-tolerant cooperative control architecture. ALLIANCE does not require any use of negotiation among robots, but rather relies upon broadcast messages from robots to announce their current activities. [3] poses solution for multiple robotic vehicles motion planning in terms of classical control theory (work is devoted to prediction of input/output reachability, structural observability, and controllability of the multiagent system). [4] describe CAMPOUT architecture for distributed control algorithms within NASA planetary surface rover systems. All sequencing in CAMPOUT is done through Finite State Machine for deterministic control. In [5] main features of cooperative grasping and transport control system for multiple quadrotors are illuminated, the main attention is paid to control signal generation for transportation of some payload simultaneous by multiple drones.

Article [6] proposes using adopted Bayesian approach for location estimation of unknown target in known, but complex environment by multiple robots. The target detection and localization is done in radio medium. Control strategy applied in [6] to solve the task is based on reduction in the uncertainty of the target estimate principle.

Previous research in cooperative map building algorithms and systems includes: [7], which considers the problem of cooperative navigation and mapping by a heterogeneous team of multiple autonomous underwater vehicles. In [7] authors address a form of cooperative Simultaneous Localization and Mapping (SLAM) in which only one vehicle is responsible for maintaining estimates of the map and poses for each robot. By combining inter-vehicle measurements with observations of the environment made by each vehicle, the result is a better knowledge of the poses of each robot in the group. In [8] describes a multi-robot map merging algorithm with the FastSLAM method and its evaluation on experimental map with landmarks. The article [9] proposes a novel extension to incremental smoothing and mapping (iSAM) algorithm, that facilitates multirobot mapping based on multiple pose graphs[10]. The article [11] describes a framework for cooperative 3D mapping of unstructured environments, which utilizing AR.Drone as UAV and combination of markerless and landmark SLAM algorithms.

To sum up above said, there exists numerous related works which are devoted to different aspects of distributed control system and cooperative map building exist nowadays. But in contrast with our work part of them describe 2D case, part of them lack of conceptual framework and architecture model, and others don't use ROS framework (or at least don't mention that in text). Yet another difference from most of the related work is usage of the low cost AR.Drone UAV which is turn poses some restriction on quality of income sensor data.

## 2   Multilevel architecture of distributed control and navigation system for quadrotors

Different approaches to design the architecture of the UAV control system exist nowadays. Some approaches suggest the usage of flat (one-level) architecture consisting of separate interacting modules tailored to solve different tasks. An example of such architecture is described in [12] and the identified tasks (and corresponding modules) are: mission planning, collaboration, contingency management, situational awareness, communications management, air vehicle management. Other approaches (more numerous) suggest splitting the functionality not only between different modules (abstracting the functionality itself) but also between different levels or layers (abstracting the level of "deliberativeness") each of which consists of bundle of modules tied together. Examples of the most known multi-leveled architectures for intelligent agents operating in real world are 3T [13], ATLANTIS [14], Aura [15] and others. Typically in robotics (as well as in UAV design) control system consists of 3 levels: deliberative (or strategic) level (top), reactive level (bottom) and intermediate (tactical) level. Top level contains modules that deal with computational expensive (AI) reasoning and planning. Intermediate level deals with analysis of spatial data and navigation tasks (e.g. SLAM, path-planning and others). Low level controls sensors and actuators (often called subsystems) of the UAV [16]. We follow that approach and decompose our control system into 3 above mentioned levels.

In our experiments we use Parrot AR.Drone platform [17] and do not override existing modules of reactive control implemented by Parrot and described in [18]. Each AR.Drone has it's own built-in flight controller which automates the execution of basic flight maneuvers: take off, land, fly forward/backward, strafe left/right, turn clockwise/counterclockwise. Commands to execute these maneuvers are sent via the open data exchange protocol from the ground station. Also we do not implement strategic level of the control system as only one high-level task, e.g. "build-map", is addressed.  As a result we mainly concentrate on the design of tactical layer capable of solving navigation tasks (e.g. identifying features and building a map of them). To do so we follow 4D/RCS reference model [19]. The main idea that lies in the basis of this model is abstract functional decomposition, e.g. each level of the architecture is composed of nodes which abstract different instances of controllable subjects and each node is composed of 4 identical functional processes (which can be viewed as bundles of modules): behavior generation, world modeling, sensory processing, value judgment (see figure 1).

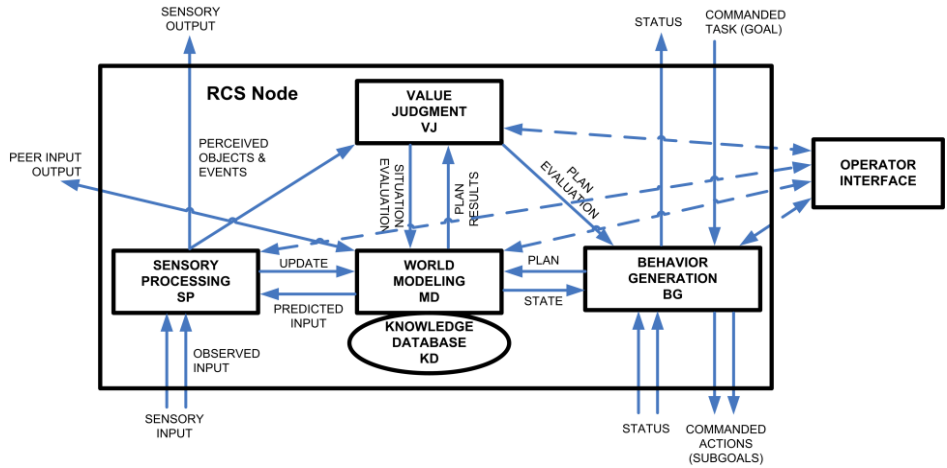

**Fig. 1.** 4D/RCS Node (from [19]).

In our case 4D/RCS processes are interpreted as following:

SP – receiving (via wi-fi link) and transcoding videostreams from two AR.Drone's cameras (forward looking and downward looking), receiving (via wi-fi link) data regarding UAV state and orientation (AT-packets contacting measurements of the internal navigation system).

VJ – AR.Drone pose estimation, recognition and identification of visual markers, computing the distances between UAV and the markers, computing position of the markers relative to local and global coordinate systems.

WM – constructing local map, integrating multiple local maps into the global (shared) one.

BG – generating rules for choosing next flight destination.

Each 4D/RCS node is in charge of controlling a single UAV. Functional processes are implemented as a bunch of Robotic operating system [20] modules (see detailed description in section 4) executed in parallel on ground station (we use standard Linux-run laptop). Global knowledge database is maintained by the system, e.g. each AR.Drone builds it's own local map which is stored on the ground station, at the same time all the gained maps are processed and a global map consistent with the measurements of all UAVs is constructed and an access to it is granted to all drones. Details of pose estimation, markers identification, map building procedures are described further as well as the details of software implementation of the system.

## 3  Map building algorithm

Implemented map building algorithm constructs a map of markers' locations as a simplified abstraction of more general map building task, that, however, can be used for practical applications where visual marker based navigation is applicable. A map is constructed simultaneously by several drones observing different subsets of mark-ers placed in corresponding parts of common environment and merging them into a global map. This is done in a following manner…

Each UAV performs a SLAM algorithm based on both inertial and visual data fused with Extended Kalman Filtering (EKF) while being in flight. During that, markers are observed and identified uniquely (they all are supposed to be unique) and their relative poses are calculated by each drone using vision. Any time a drone observes a marker, it checks it's existence on the global map which storage is shared appropriately. Locations of markers that appear to be already known are used to correct current UAV position state. New markers are added to the global map and their positions are fused with the next few observations to reduce inaccuracy of visual pose measuring. As subsets of markers visible by different UAV's may not overlap for a long time, these markers are stored linked to independent coordinate frames until some marker or a group is observed by two or more drones, then coordinate systems are merged, and translation of markers' poses from eliminated coordinate frame to conserved one is performed. As markers' total count is assumed to be relatively small, bundle adjustment from a set of keyposes like in [21] is used to unbias the map.

We utilize ArUco [22] marker system and library capable of detection, recognition and 6-DOF pose estimation of up to 1024 different bar-codes, distinctive reliably with the help of modified Hamming code. With the camera calibrated and known marker parameters it becomes easy to build the map in meter units.

We use Extended Kalman Filtering for drone pose estimation with state vector $h_t = (x_t, y_t, z_t, \alpha_t, \beta_t, \gamma_t)^T$, where elements correspond to 3 spatial coordinates and 3 Euler angles in global frame (totally global or one of the independent global ones) correspondingly. Odometry observation and prediction models are similar to that described in [23]. Pose observation by means of vision relies on relative pose estimation of all visible markers with known positions at some moment. Given observed marker pose, described as translation vector $T_m$ and rotation matrix $R_m$ in camera coordinate system, we first calculate approximate measurement noise covariance matrix (assuming measured values along 3 axes independent for simplicity). Then we use observed relative marker coordinates to calculate UAV's global pose observation; the covariance is combined with the covariance assigned to the marker on the map. The EKF state is then updated using obtained pose observation and resulting covariance. The EKF state is updated with the information obtained from each of observed already known markers.

Merging coordinate systems of drones when they observe the same markers for the first time is done in the following manner. Firstly RT matrix is calculation as 3D affine estimation based on matching markers, then this matrix is used to transform markers poses from one drone coordinate system to other drone coordinate system. After that if a new matching markers have found, RT matrix is refined and markers poses are transformed again.

## 4 Software implementation of distributed control and navigation system for quadrotors based on ROS

Software implementation of described system is based on ROS framework and 4D/RCS scheme. Structure of software is shown on figure 2.

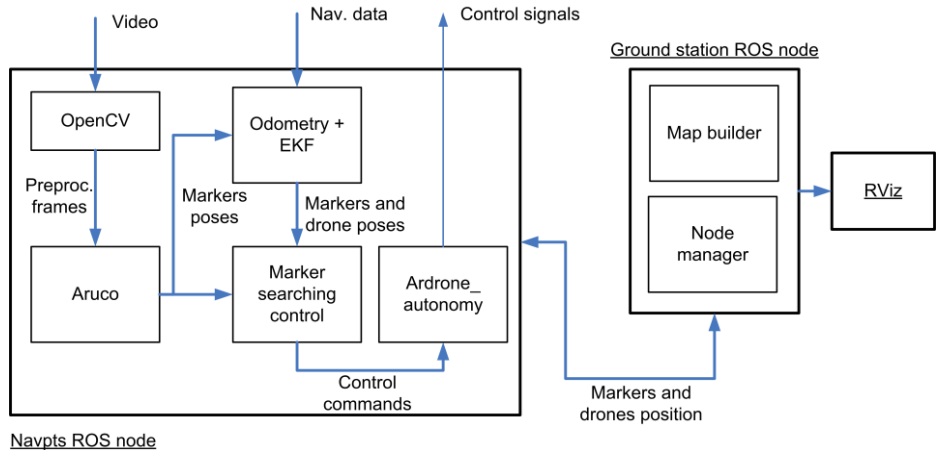

**Fig. 2.** – Structure of distributed control system software implementation based on ROS

ROS packages, nodes and also external libraries are shown on figure 2. There are two main ROS nodes: "Navpts" node and "Ground station" node. Several instances of Navpts node are executed simultaneously each of which controls a single AR.Drone. Ground station node implements map building and sharing algorithm and only one instance of that node is executed at a time. If we describe Navpts node following 4D/RCS we could say that OpenCV library relates to sensory processing, Aruco and EKF packages to value judgment, marker searching control package to behavior generation. 4D/RCS process of world modeling process relates to distributed between Navpts and Ground station nodes map representation.

Ground station node consists of map builder module, which implement described algorithms, and node manager, that start and remap necessary topics for Navpts nodes.

For map visualization we use Rviz package of ROS.

## 5 Evaluation and conclusions

We evaluate our system implementation in laboratory environment. The result is shown in form of merged map on figure 3 (right).

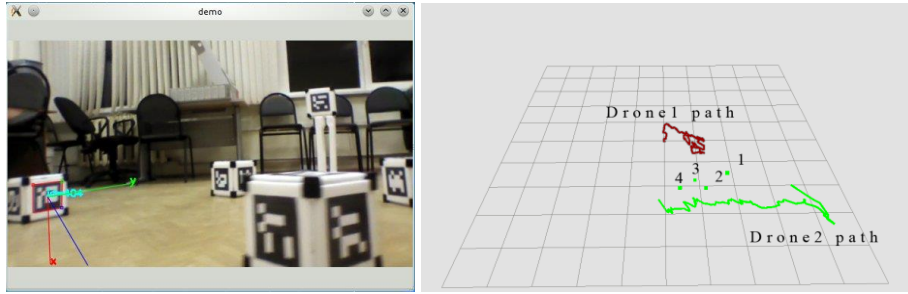

**Fig. 3.** Marker detection on image (left) and merged map in Rviz (right)

On figure 3 (left) experimental setup and detection of one marker is shown. On figure 3 (right) shared map is shown and trajectories of drones and position of common markers are indicated.

As a conclusion we could point out, that our implementation of map building isn't optimal solution for such task compared to related works and we use it like an illustration of chosen architecture and software implementation of distributed control and navigation system for quadrotors in GPS-denied environments. We plan to extend our system in the following direction: integrate with more powerful SLAM algorithm, which will be applicable in environment with unknown landmarks, and modify behavior generation process by introduction cooperative planning to struggle with situation in which one of drones couldn't detect any markers and to increase efficiently of the system.

**Acknowledgments.** This work was supported by the Ministry of Education and Science of the Russian Federation (№ 14.577.21.0030 agreement for a grant on "Conducting applied research for the development of intelligent technology and software systems, navigation and control of mobile technical equipment using machine vision techniques and high-performance distributed computing").